\pdfoutput=1

\documentclass[11pt]{article}

\usepackage[review]{acl}

\usepackage{times}
\usepackage{latexsym}
\usepackage{graphicx}
\usepackage{multirow}
\usepackage{booktabs}
\usepackage{tabularx,tabulary}
\usepackage{caption}
\usepackage{xcolor}
\usepackage{soul}
\usepackage{adjustbox}      
\definecolor{ForestGreen}{RGB}{34,139,34}
\definecolor{plum}{RGB}{142, 69, 133}
\definecolor{Bittersweet}{RGB}{254, 111, 94}
\definecolor{BurntOrange}{RGB}{204, 85, 0}

\usepackage[T1]{fontenc}

\usepackage[utf8]{inputenc}
\usepackage{todonotes}

\usepackage{microtype}

\title{DEAM: \underline{D}ialogue Coherence \underline{E}valuation using \underline{A}MR-based Semantic \underline{M}anipulations}

\author{
Sarik Ghazarian,\textsuperscript{\rm 1}
Nuan Wen,\textsuperscript{\rm 1} 
\textbf{Aram Galstyan},\textsuperscript{\rm 1}
\textbf{Nanyun Peng}\textsuperscript{\rm 1, 2} \\
\textsuperscript{\rm 1}University of Southern California / Information Sciences Institute \\
\textsuperscript{\rm 2}Computer Science Department of University of California, Los Angeles \\
\{sarik, nuanwen, galstyan\}@isi.edu,
violetpeng@cs.ucla.edu
}

\begin{document}
\nolinenumbers
\maketitle
\begin{abstract}

Automatic evaluation metrics are essential for the rapid development of open-domain dialogue systems as they facilitate hyper-parameter tuning and comparison between models. Although recently proposed trainable conversation-level metrics have shown encouraging results, the quality of the metrics is strongly dependent on the quality of training data. 
Prior works mainly resort to heuristic text-level manipulations (e.g. utterances shuffling) to bootstrap incoherent conversations (negative examples) from coherent dialogues (positive examples). Such approaches are insufficient to appropriately reflect the incoherence that occurs in interactions between advanced dialogue models and humans. 
To tackle this problem, we propose DEAM, a \underline{D}ialogue coherence \underline{E}valuation metric that relies on \underline{A}bstract Meaning Representation (AMR) to apply semantic-level \underline{M}anipulations for incoherent (negative) data generation. AMRs naturally facilitate the injection of various types of incoherence sources, such as coreference inconsistency, irrelevancy, contradictions, and decrease engagement, at the semantic level, thus resulting in more natural incoherent samples. Our experiments show that DEAM achieves higher correlations with human judgments compared to baseline methods on several dialog datasets by significant margins. We also show that DEAM can distinguish between coherent and incoherent dialogues generated by baseline manipulations, whereas those baseline models cannot detect incoherent examples generated by DEAM. Our results demonstrate the potential of AMR-based semantic manipulations for natural negative example generation. 

\end{abstract}

\section{Introduction} \label{section:intro}

\begin{figure}[t]
\centering
\includegraphics[width=\linewidth]{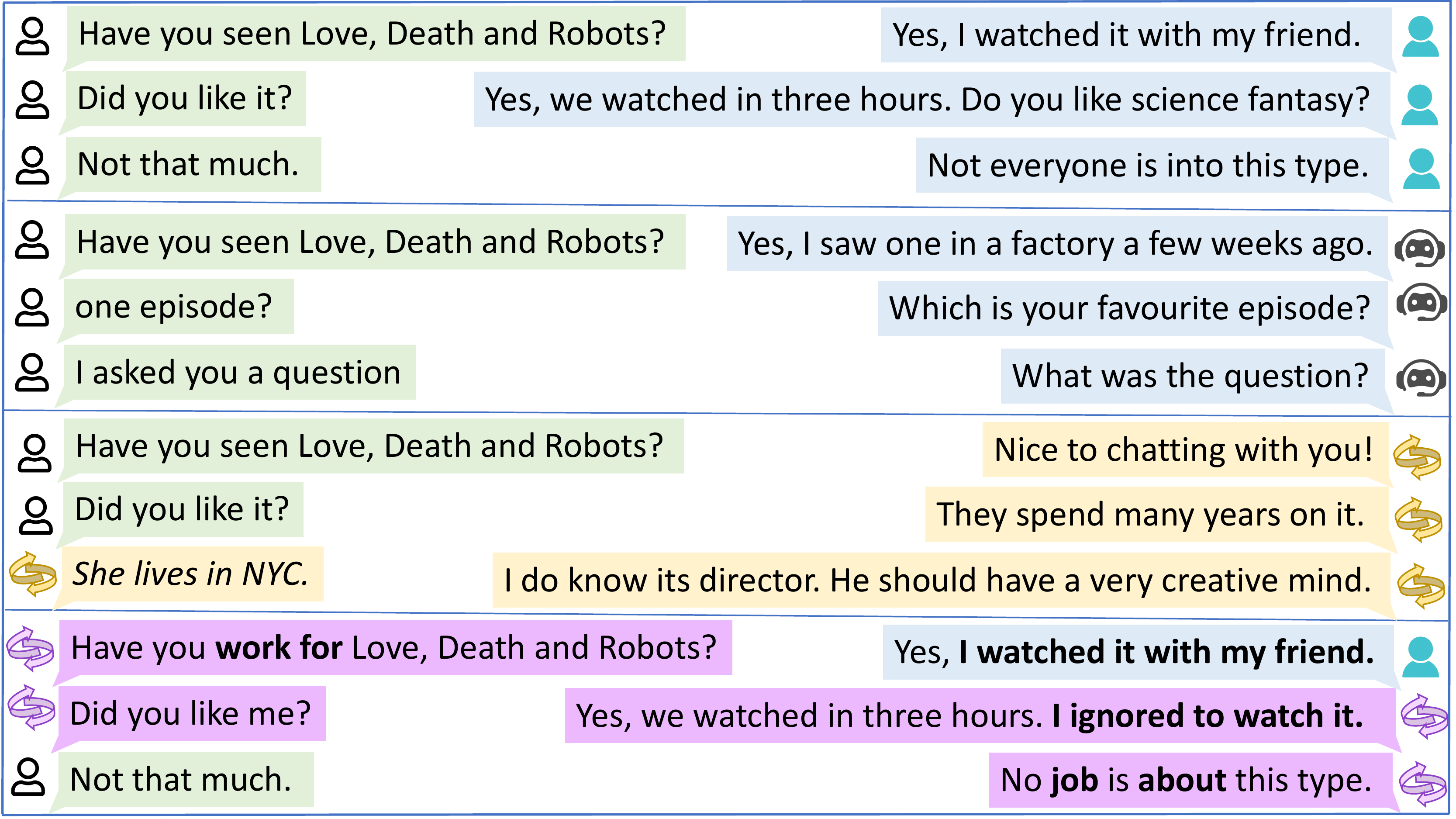}
\caption{Examples of human-human (first dialogue) and human-chatbot (second dialogue) conversations alongside manipulations resulted from baseline (indicated with yellow color) and our proposed perturbations (indicated with purple color), respectively. Similar to the human-chatbot interaction, our manipulations result in more subtly-incoherent dialogue compared to baseline manipulations.}
\label{fig:perts_exmp}
\vspace{-1.5em}
\end{figure}

Despite the effectiveness of large pretrained language models~\cite{radford2019language, lewis2020bart} for dialogue response generation~\cite{zhang2020dialogpt, adiwardana2020meena, ghazarian2021discol}, it is still challenging for the models to imitate human-human conversations and maintain conversational-level coherence. To better evaluate such models, recent works propose trainable automatic evaluation metrics to benchmark and compare the performance of dialogue models~\cite{wu2020diverse, zheng2021dynaeval}. 
Most trainable automatic evaluation metrics focus on turn-level interactions, where they learn to assess the quality of one user-system utterance pair~\cite{tao2018ruber, huang2020grade, ghazarian2020engagement}.  
However, these metrics cannot appropriately model the whole conversation flow~\cite{yeh2021comprehensive}, and thus are insufficient for dialogue-level evaluation. 

In this work, we focus on the automatic evaluation of the coherence of \textit{dialogues}, which is under-explored.
\textbf{Coherence} is a conversation-level metric that measures how well the utterances in a conversation are unified leading to a consistent interaction~\cite{byron1998apreliminary, mesgar2020dialogue}. 

Previous works pursue different models such as graph-based~\cite{vakulenko2018measuring, zheng2021dynaeval} or text-based~\cite{mesgar2020dialogue} approaches to develop automatic trainable coherence evaluation metrics. Those models take a \textit{contrastive learning} approach, where they build binary classifiers to differentiate positive, or coherent examples from negative, or incoherent dialogues. Those classifiers are usually trained on datasets constructed by using human-human conversations as positive examples and applying \textit{text-level heuristic manipulations} to generate incoherent conversations. The \textit{text-level manipulations} directly change the structures of the conversation such as shuffling the order of utterances, replacing some random utterances from external conversations~\cite{vakulenko2018measuring, mesgar2020dialogue, zheng2021dynaeval}, as shown in the third dialogue of Figure~\ref{fig:perts_exmp}.

We posit that such text-level manipulations are too simplistic to adequately represent more nuanced coherence errors presented in the current state-of-the-art dialogue systems. For example, the second conversation in Figure~\ref{fig:perts_exmp} shows a human-system interaction from the FED dataset~\cite{mehri2020fed}, where the incoherence is much more subtle than the ones created by text-level manipulations.

In this paper, we investigate \textit{manipulation techniques} to generate negative samples that represent coherence errors more likely to happen in the state-of-the-art dialogue systems. To this end, we propose DEAM~\footnote{Our proposed manipulations, data, and trained models can be found at \url{https://github.com/PlusLabNLP/DEAM}}, a model that uses Abstract Meaning Representation (AMR) to apply \textit{semantic-level manipulations} to generate negative examples. 
AMRs are intended to capture the meaning of a sentence by abstracting away irrelevant syntactic features. Thus, injecting targeted and controlled perturbations into an AMR is easy and can introduce semantic incoherence into the corresponding sentences. 

DEAM starts with parsing conversations into semantic AMR representations and then injects incoherence types that are usually observed in current state-of-the-art models into the AMR graphs. It concludes this process by translating the manipulated AMRs back to conversations as negative examples using a controllable generation model. 
A fine-tuned RoBERTa model is then trained on the created dataset to distinguish coherent and incoherent conversations as the evaluation metric. 

Our main contributions are as follows:

\begin{itemize}
\setlength\itemsep{-.2em}
\item We propose DEAM, an evaluation metric that leverages AMR graphs and injects incoherence sources at the semantic level to generate incoherent conversations for training. 
\item We propose four manipulation strategies to represent four common incoherence sources of the state-of-the-art dialogue models: contradiction, coreference inconsistency, irrelevancy and decrease engagements. 
\item We empirically show that the model trained on our proposed manipulations significantly outperforms strong baselines in terms of correlation with human judgments. 
Moreover, DEAM is capable of distinguishing positive and negative examples generated by baselines that use text-level manipulations, whereas the opposite is not true -- classifiers trained on text-level manipulations cannot detect negative examples generated by DEAM. This demonstrates the effectiveness of the semantic-level AMR-based manipulations.
\end{itemize}

\section{Related Works} \label{related_works}

Automatic evaluation of open-domain dialogue systems has a multifaceted nature with many fine-grained quality aspects~\cite{mehri2020fed}. Turn-level aspects show the quality of the system's utterance given a dialogue context from different perspectives including appropriateness, relevance, engagement, and etc~\cite{lowe2017adem, tao2018ruber, ghazarian2020engagement}. Whereas, conversation-level facets such as coherence, diversity, informativeness take into account the whole dialog flow~\cite{vakulenko2018measuring, zheng2021dynaeval, mehri2020fed}. 

Dialogue coherence evaluation is pertinent to discourse coherence since a dialogue is counted as a multi-party discourse. Similar to discourse coherence, many original coherence evaluation metrics derived from the Centering Model for monitoring the local focus of utterances and their entities distribution~\cite{grosz1986discourse, miltsakaki2004essay, lapata2005entitygrid}. A group of studies assess the coherence of dialogues with respect to entities and dialogue acts~\cite{cervone2020coherent, mesgar2020dialogue}. Another inspected approach for dialogue coherence evaluation is to represent dialogue in a structured graph format where contextually dependent neighbor utterances or concepts are connected nodes in the graph~\cite{vakulenko2018measuring, mesgar2020dialogue, huang2020grade}. Graph convolutional networks are used to complete this task. 

High-quality training dataset is identified as one of the momentous and indelible components in automatic coherence evaluation. Some previous works construct such datasets by collecting human judgments~\cite{higashinaka2014evaluate, cervone2020coherent}. While many recent works rely on a more timely and costly affordable approach by automatically generating negative samples. The utterances of the coherent conversations from human-human interactions are manipulated by shuffling their order, inserting or replacing irrelevant utterances~\cite{vakulenko2018measuring, mesgar2020dialogue, huang2020grade, zheng2021dynaeval}. In this work, we show that such changes can not truly represent machine-generated incoherent conversations.
One work that is closely related to us and proposed abstract-level manipulations is \newcite{ghazarian2021plot}. However, their application domain is open-domain story evaluation, and they use story plot, rather than AMR, for the manipulation, which is more domain-specific.

\section{DEAM Overview} 
Our goal is to build an evaluation metric that measures the conversation-level coherence of dialogues. 
We follow the trainable evaluation metrics~\cite{vakulenko2018measuring} to formulate the evaluation as a classification task. We train the evaluator on positive (coherent) and negative (incoherent) conversations, and take the predicted probability for the positive class as the coherence score. 

\begin{figure}[t]
\centering
\includegraphics[width=\linewidth,height=.5\linewidth]{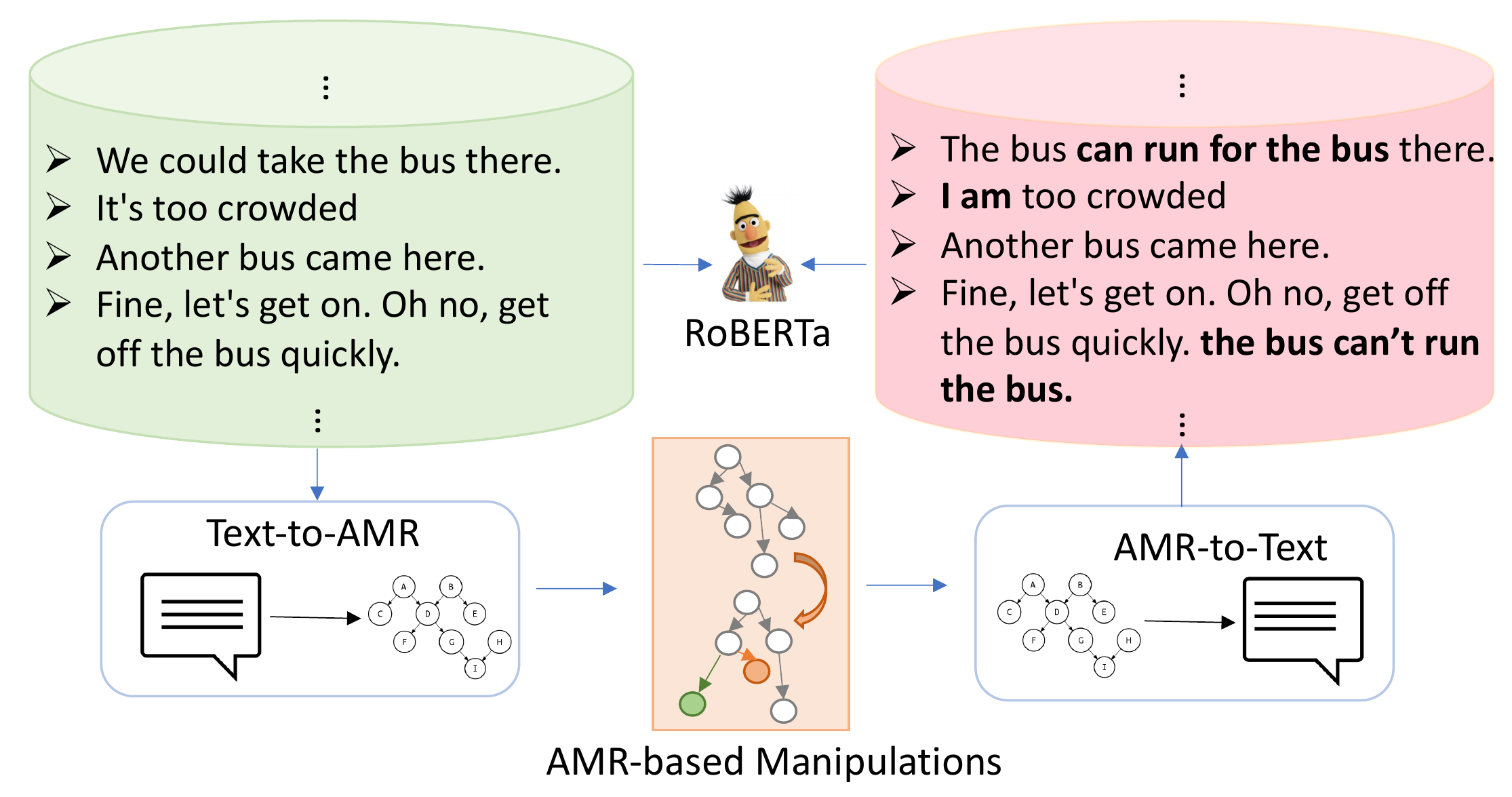}
\vspace{-.5em}
\caption{Overall architecture of DEAM metric trained on positive (green box) interactions and negative (red box) conversations generated from AMR-based manipulations (orange box)}
\label{architecture}
\vspace{-1.2em}
\end{figure}

As is discussed above, the main challenge for building a reliable metric is to obtain negative samples that can adequately represent the incoherence issues presented in advanced dialogue systems. 
To this end, we propose to generate negative examples by leveraging AMR-based manipulations. 
We then build a RoBERTa-based classifier as the evaluation metric by fine-tuning RoBERTa on the automatically generated training data. 
Figure~\ref{architecture} illustrates an overview of our proposed evaluation method. 

The first step of DEAM is to apply Text-to-AMR models to the conversations. Text-to-AMR or AMR parsing~\cite{jin2019amr, xu2020amr, zhou2021amr, lam2021amr} that translates conversation texts to directed and acyclic AMR graphs containing relation edges between concept nodes~\cite{banarescu2013amr} has been effectively accomplished by transformer-based models in a sequence-to-sequence (seq2seq) training fashion~\cite{xu2020amr, zhou2021amr}. We use the fine-tuned T5~\cite{raffel2020t5} model\footnote{We leverage the released parse\_t5 model from \url{https://github.com/bjascob/amrlib}} for this purpose.

\begin{figure}[t]
\centering
\includegraphics[width=\linewidth]{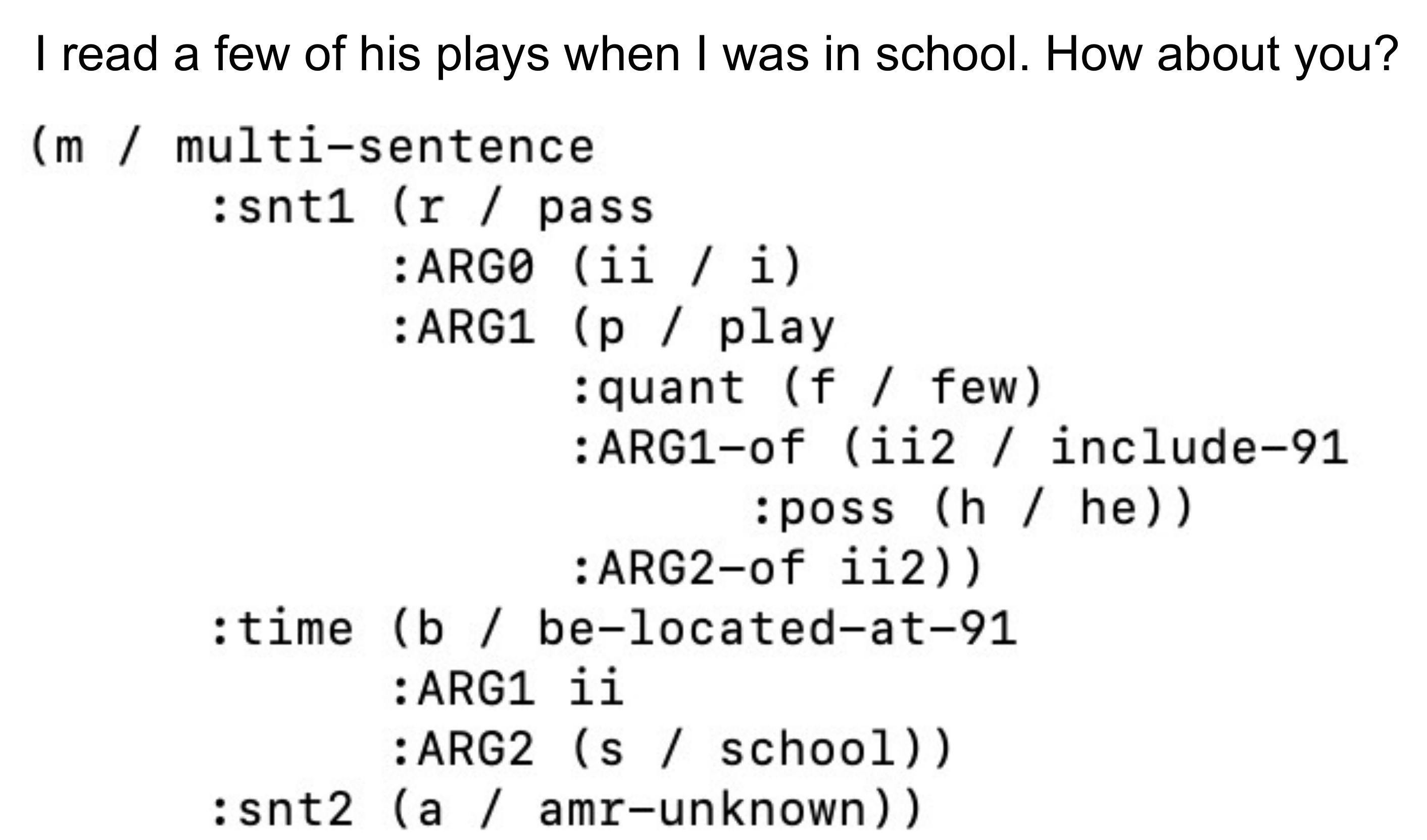}
\vspace{-.5em}
\caption{AMR representation of a dialogue utterance}
\label{amr_graph}
\vspace{-1em}
\end{figure}

We then manipulate the AMR graphs (section~\ref{amr_mplts}) and back translate them into conversation texts to be used as negative examples for training the text-based coherence evaluator.
Similar to AMR parsing, we use finetuned T5\footnote{We leverage the released generate\_t5wtense model from \url{https://github.com/bjascob/amrlib}.} that is shown to be effective for the AMR-to-Text generation task~\cite{mager2020amr, ribeiro2020amrtotext}.

\section{Incoherent Dialogue Generation} \label{section:mplts}
The challenge that automatic trainable evaluation metrics face is in providing training data that can appropriately replicate moderate to low quality conversations with incoherence sources that usually happen in the current dialogue models. 
The common solution is to apply manipulations to positive conversations. In this section, we summarize the baselines manipulations and state our proposed AMR-based perturbations.

\subsection{Baselines Manipulations} \label{baseline_mplts}
Baseline manipulations can be classified as:

1) Shuffling-based manipulations: In such manipulations, turns order~\cite{vakulenko2018measuring}, sequence of speakers utterances~\cite{mesgar2020dialogue, vakulenko2018measuring, zheng2021dynaeval}, or the position of the first and second sections of conversations~\cite{vakulenko2018measuring} are swapped. 

2) Insertion-based manipulations: This group of manipulations add incoherence sources by replacing~\cite{mesgar2020dialogue, zheng2021dynaeval} or inserting~\cite{mesgar2020dialogue} a random utterance from a randomly selected conversation. Each baseline metric fuses multiple manipulations, hence we use their citations \cite{vakulenko2018measuring}, \cite{mesgar2020dialogue} to easily refer them in later sections.

\subsection{AMR-based Manipulations} \label{amr_mplts}
AMR is originally proposed by~\citet{banarescu2013amr} as a semantic representation language that helps to abstract away the text from surface syntactic. 
Many abstract-level semantic information such as named entities, negations, questions, coreferences and modalities in the texts can be encoded by AMR graphs. These potential capabilities of AMR make it lucrative in many semantic-related NLP tasks such as summarization~\cite{liao2018sum} and machine translation~\cite{song2019amr}. Conversations between two interlocutors contain many semantic details that can be captured by these graphs. Therefore, we explore AMR features' usage in the dialogue systems evaluation task by manipulating the AMR graphs of coherent conversations, each manipulation reflecting a specific reason for incoherence in dialogue systems.  
Figure~\ref{amr_graph} demonstrates a linearized version of an utterance AMR graph.

In AMR graphs, entities and concepts are shown as nodes and their relations are depicted with various relation edges ~\cite{banarescu2013amr}. Each AMR concept is either a word, or a PropBank framesets keyword~\cite{kingsbury2002propbank}. The PropBank framesets with their predefined arguments are used to abstract away concepts from syntactic structures. As an example, \textit{located} concept of PropBank framesets in Figure~\ref{amr_graph} comes with two arguments the subject (\textit{i}) and the place (\textit{school}).  

In DEAM, we pursue the idea of manipulating abstract-level semantic information extracted from AMRs to generate incoherent conversations.
In this work, we mainly focus on four major logical flaws that could happen in state-of-the-art dialogue models such as cases when a chatbot contradicts its previously stated utterances, uses incorrect coreferences, responds users with irrelevant utterances, does not engage enough in the conversation. We explain each of these logical flaws in detail. 

\subsubsection{Contradiction}
\begin{figure*}[t]
\centering
\includegraphics[width=.88\linewidth,height=.4\linewidth]{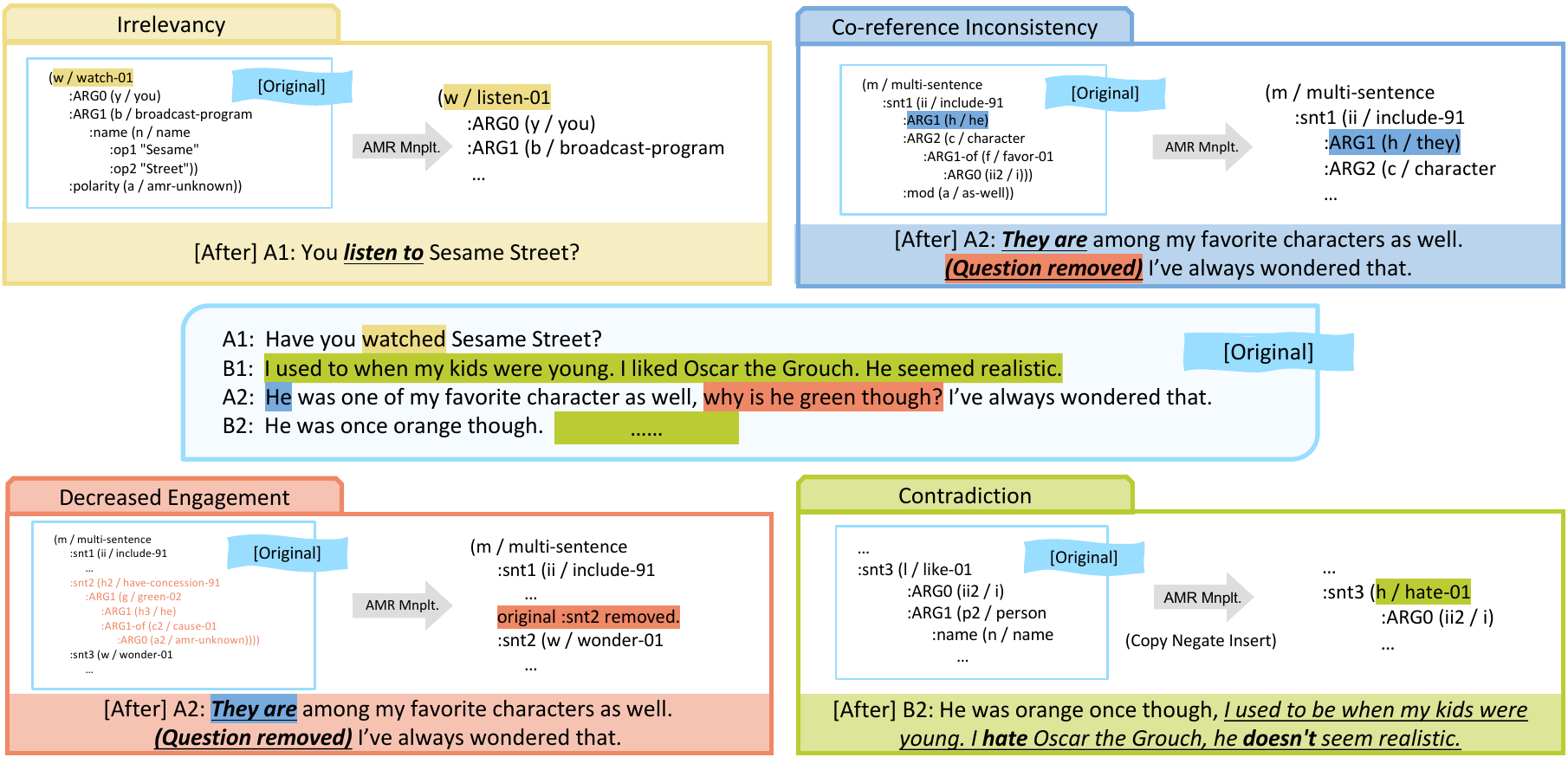}
\vspace{-.5em}
\caption{An abbreviated sample conversation to illustrate four different AMR-based DEAM manipulations}
\label{manipul_plt_fig}
\end{figure*}

One of the common issues that dialogue systems struggle with is directly or indirectly contradicting previous utterances in dialogue. To replicate this type of error, a contradicted version of a subgraph from the original AMR is copied to other locations. This negative form AMRs can be accomplished by directly adding \textit{polarity} to the concepts or replacing concepts with their antonyms that hold \textit{Antonym}, \textit{NotDesires}, \textit{NotCapableOf}, and \textit{NotHasProperty} relations in ConceptNet~\cite{speer2012conceptnet}. After adding contradictions, the AMR-to-Text model will use the encoded context to output incoherent yet natural conversations. In the bottom right example of Figure~\ref{manipul_plt_fig}, speaker B contradicts its previously stated opinion that badly effects the linkage of the utterances.

\subsubsection{Coreference Inconsistency}
The coherence of a conversation is preserved by the correct references of previously mentioned entities and words in the dialogue context. Pronouns in the conversation play an essential role in this regard. 
Coreferences in AMRs are presented as arguments (\textit{ARG}) and all three different types of pronouns such as subjective, objective and possessive pronouns are shown in their subjective format.

To disrupt the coreferences relations, we randomly replace some pronouns in the conversation's AMR with another pronoun or noun identified as \textit{ARG} or operand (\textit{op}) from the same conversation. After replacements, the AMR-to-Text model adapts other sections of the utterance accordingly and reassures us that outputs have natural look and correct grammar.
The third utterance in Figure~\ref{manipul_plt_fig} demonstrates an example of coherence inconsistency which makes the utterance to be not logical.  

\subsubsection{Irrelevancy}
Random utterance substitution from other conversations is a simple way to inject incoherence sources in dialogues, which has been frequently used in prior work~\cite{tao2018ruber, ghazarian2018bertruber, mesgar2020dialogue, zheng2021dynaeval}. Conversations with completely off-topic utterances are rarely generated by advanced dialogue models due to their ability in encoding dialogue history for continuing the conversation.

We propose to apply irrelevancy sources to AMR graphs. We select some AMR items such as concepts, \textit{op}s, \textit{ARG}s and replace them with random items from other utterances. In this approach, the replacement items are not from randomly selected conversations but still, they do not fit well in their newly selected locations which hurts the coherence of the conversation. In Figure~\ref{manipul_plt_fig} \textit{watch} is replaced with \textit{listen}. The benefits of using AMR-to-Text model emerge here where some new adaptations (such as \textit{to}) have been augmented with new verb replacement to give the utterance a fluent look.

\subsubsection{Decrease Engagement}
In coherent conversations, speakers exchange opinions about different topics by stating detailed information, asking and answering questions. This coherence will be faded if one of the interlocutors evades to answer questions or talk in detail.
In contrast to previous works that ignored this important feature, we augment such kind of incoherence sources into the negative sampling generation. In order to decrease the engagement of coherent conversations, we take the advantage of AMRs which are able to demonstrate detailed utterances and those containing questions.
In AMR graphs, detailed utterances include more number of nested layers and concepts, \textit{ARG}s and \textit{op}s. Question-type utterances can be easily distinguished via \textit{amr-unknown} concept notation and therefore with relying on AMRs the goal of decreasing engagement in the conversation is easily achievable.

We propose three different approaches to decrease the engagement and consequently the coherence of the conversations:

1) Remove question-type utterances in the conversation: we select a multi-sentence utterance including \textit{amr-unknown} concept and remove it and all its children nodes from the graph; 2) Remove the most detailed utterance in the conversation: the utterance having the largest depth in the graph is selected as the utterance with the most transferred information and all its children alongside its parent concept are removed from the graph; 3) Remove fine-grained information in the utterances: the main concepts' detailed information that are presented as \textit{ARG} or \textit{op} in the AMRs are randomly selected and eliminated from the graph. The higher-level concepts in the graph are preserved while its lower-level child nodes are deleted which makes the utterance not transfer its meaning and diminishes the linkage of topics. The question part in the third utterance of Figure~\ref{manipul_plt_fig} has been removed causing the coming utterances to not be completely sensible.

\section{Experimental Setup} \label{section:exp_setup}

We compare DEAM and its negative example generation techniques with baseline models and manipulations. We aim to have a data-driven analysis under three setups: 

\textbf{Setup 1)}: In this setup, we compare DEAM with baseline models by varying both the data manipulation strategies and the classification models. We fix the positive examples to be the same set of human-human conversations.

\textbf{Setup 2)}: Since baseline models are trained on different datasets, we conduct pairwise comparisons between DEAM and each baseline evaluator by training on the baseline's dataset. Note that we only take the positive examples from the baseline's dataset, and apply different manipulations to get negative examples to compose a balanced set for training. We also train different classifiers (DEAM vs. baselines) for the evaluators. 

\textbf{Setup 3)}: This setup is designed to show the effectiveness of different manipulations to generate negative examples. We fix the positive examples and the classifier (i.e., RoBERTa). 

\subsection{Datasets}
\subsubsection{Training Datasets}
We conduct our experiments on two crowd-sourced datasets, TopicalChat~\cite{gopalakrishnan2019topical} and PersonaChat~\cite{zhang2018personachat}.
Both datasets are composed of conversations between Amazon Mechanical Turk (AMT) participants.
In TopicalChat, AMT workers were supposed to have coherent and engaging conversations regarding the provided reading sets about different topics, while in the PersonaChat dataset coherent conversations were conditioned on the provided 1155 personas each including 5 personality description sentences collected via AMT.
We take these conversations as coherent conversations. We follow DEAM's steps to generate and add a balanced number of incoherent conversations. Table \ref{data_stats} shows the  train/valid statistics of the newly constructed datasets called \textsc{Topical\_DEAM} and \textsc{Persona\_DEAM}. 
\begin{table}
\centering
\small
\begin{tabular}{lccc}
\toprule
\textbf{Dataset} & \textbf{size} & \textbf{conv. len} & \textbf{utt. len}\\
\midrule
\textsc{\textsc{Topical\_DEAM}} & 17.3k/2.2k & 530/530 & 24/24\\
\textsc{\textsc{Persona\_DEAM}} & 17.9k/2.0k & 187/202 & 13/13\\
\textsc{\textsc{FED}} & 125 & 168 & 11\\
\textsc{\textsc{DSTC9}} & 2.2K & 318 & 11\\
\bottomrule
\end{tabular}
\caption{Statistics (size, average length of conversations and utterances) of TopicalChat and PersonaChat train/valid datasets (augmented with AMR-based manipulated conversations), alongside with FED and DSTC9 test datasets.}
\label{data_stats}
\vspace{-1.5em}
\end{table}

\subsubsection{Evaluation Datasets}

In the literature of automatic evaluation metrics, the prevalent way of assessing the evaluators' performance is to compare the correlation of their predicted scores with human judgments. FED~\cite{mehri2020fed} and Interactive Evaluation of Dialog track of the Dialog State Tracking Challenge 9 (DSTC9)~\cite{gunasekara2020overview}  are two publically available benchmark datasets including human ratings on the coherence aspect of human-human or human-systems conversations. 

The participants of FED dataset, have judged 125 conversations; 41 human-human, 44 human-Mitsuku chatbot and 40 human-Meena~\cite{adiwardana2020meena} chatbot. Humans have assessed the conversations from 11 conversation-level evaluation aspects including the coherence and overall scores. Each conversation in FED is judged by 5 distinct annotators. Coherence and overall scores are in the range of 0-2 and 0-4, respectively. 

In DSTC9 dataset\footnote{\url{https://github.com/exe1023/DialEvalMetrics}}, AMT workers have rated 2200 conversations between invited participants and knowledge-grounded response generation models using the same 11 fine-grained conversation-level evaluation aspects. Coherence and overall scores that we use in our experiments are in the range of 1-3 and 1-5, respectively. In our experiments, we take the average of judgments for conversations with more than one annotator's ratings and compute the Spearman correlations between human evaluations and evaluator's generated scores.

\subsection{Implementation Details} 
In our work, we train and run all the models on a machine with a GeForce RTX 2080 Ti GPU. 
We fine-tune RoBERTa-large pretrained model on \textsc{Topical\_DEAM} and  \textsc{Persona\_DEAM} datasets for three epochs and optimize parameters using Adam optimizer with 1e-5 learning rate. 

To conduct experiments in setup 1, we train \citet{vakulenko2018measuring}'s graph-based model for 128 epochs with 1e-5 learning rate. \citet{mesgar2020dialogue}'s LSTM-based model is trained for 8 epochs with 5e-5 learning rate. We retrain DynaEval~\cite{zheng2021dynaeval} for 20 epochs. 
All baselines are trained using Adam optimizer.
Due to not publically published models proposed by ~\citet{vakulenko2018measuring} and \citet{mesgar2020dialogue}, we need to retrain these models on their original datasets; Ubuntu~\cite{lowe2015ubuntu} and DailyDialog~\cite{li2017dailydialog}; using the same hyperparameters published in the aforementioned papers to complete experiments in setup 2. We use DynaEval's published checkpoints to run experiments in this setup. 

In experimental setup 3, we start from \textsc{TopicalChat} and \textsc{PersonaChat} datasets, and augment negative samples pursuing different manipulation techniques. We fix the evaluator and finetune RoBERTa-large model for 3 epochs with a 1e-5 learning rate. 
Since ~\citet{vakulenko2018measuring}'s proposed manipulations are in the entity level,  we adapt the perturbations to the text level by replacing the sequence of entities with a sequence of utterances substitutions to be acceptable by the RoBERTa model.

\section{Results} \label{results}
Through our experiments, we report the Spearman correlation of evaluation metrics with human annotations under different experimental setups. 

\subsection{Metrics Performance}
Table~\ref{model_comp} depicts the quantitative results for different evaluation models on both FED and DSTC9 datasets of experimental setup 1. According to the reported correlations, the superiority of DEAM shown in the last row versus other baselines is obviously recognizable. This superiority could originate from the subtle negative sampling technique.

In experimental setup 1, manipulation techniques and models vary between evaluators, therefore we complete our investigation via experimental setup 2 by conducting one by one comparison of DEAM with each baseline model training each pair on the same dataset that the baseline model has been trained on. Table~\ref{pair_model_comp} shows the output of this type of pairwise comparisons separated each pair into one section. Even though most of the baseline models correlations increased, yet DEAM takes the lead. It is noteworthy that the correlation of \textsc{DynaEval} reported in the original paper for FED dataset has decreased, this could be due to the less number of negative samples that we consider for each positive conversation and its major impact on this model's performance. 

\begin{table}[t]
\begin{center}
\small
\begin{tabular}{lcccc}
\toprule
 \multirow{2}{*}{\textbf{Model}} & \multicolumn{2}{c}{\bf FED} & \multicolumn{2}{c}{\bf DSTC9} \\ 
 & Coh & Ovrl. & Coh & Ovrl. \\
\midrule
  \citet{mesgar2020dialogue}
   & 0.10 & -0.01 & 0.02 & 0.05\\ 
\midrule 
  \citet{vakulenko2018measuring}
   & 0.13 & 0.10 & -0.001 & -9.6e-5 \\ 
\midrule
  \textsc{DynaEval}  
    & -0.36  & -0.4  & -0.03 & -0.01 \\
\midrule
  \textsc{DEAM}
     & \bf 0.47 &  \bf 0.55  &  \bf 0.19 & \bf 0.20 \\
  \bottomrule 
\end{tabular}
\end{center}
\caption{Spearman Correlations of different models with human judgements trained on \textsc{TopicalChat} and \textsc{PersonaChat} datasets following different manipulations for negative sample generation (setup 1).}
\label{model_comp}
\end{table} 

\begin{table}[t]
\begin{center}
\small
\begin{tabular}{lcccc}
\toprule
 \multirow{2}{*}{\textbf{Manipulation}} & \multicolumn{2}{c}{\bf FED} & \multicolumn{2}{c}{\bf DSTC9} \\ 
 & Coh & Ovrl. & Coh & Ovrl. \\
\midrule
  \citet{mesgar2020dialogue}     
   & 0.29 & 0.24 & 0.15 & 0.14\\ 
\midrule 
  \citet{vakulenko2018measuring} 
   & 0.29 & 0.20 & 0.15 & 0.14 \\ 
\midrule
  \textsc{DynaEval}  
    & 0.32  & 0.25  & 0.14 & 0.15 \\
\midrule
  \textsc{DEAM}
     & \bf 0.47 &  \bf 0.55  &  \bf 0.19 & \bf 0.20 \\
  \bottomrule 
\end{tabular}
\end{center}
\caption{Spearman Correlations of the same RoBERTa-large models finetuned on \textsc{TopicalChat} and \textsc{PersonaChat} datasets augmented with incoherent conversations generated by different manipulation techniques (setup 3).}
\label{mplts_comp_table}
\vspace{-.5em}
\end{table}

\begin{table*}[t]
\begin{center}
\small
\begin{tabular}{lllcccc}
\toprule
 \multirow{2}{*}{\textbf{Model}} & \multirow{2}{*}{\textbf{Manipulations}} & \multirow{2}{*}{\textbf{Dataset}} & \multicolumn{2}{c}{\bf FED} & \multicolumn{2}{c}{\bf DSTC9} \\ 
 & & & Coherence & Overall & Coherence & Overall \\
\midrule
 \citet{vakulenko2018measuring} & \citet{vakulenko2018measuring} & \textsc{Ubuntu} & 0.17  & 0.15  & -0.04 & -0.1 \\
 \textsc{DEAM} & \textsc{DEAM} & \textsc{Ubuntu} & \bf 0.18  & \bf 0.25  & \bf 0.16 & \bf 0.15 \\
\midrule
\citet{mesgar2020dialogue} & \citet{mesgar2020dialogue} & \textsc{DailyDialog} & -0.36 & -0.47 & 0.13 & 0.14\\ 
  \textsc{DEAM} & \textsc{DEAM} & \textsc{DailyDialog} & \bf 0.34 & \bf 0.36 & \bf 0.17 & \bf 0.18 \\
\midrule
 \textsc{DynaEval} & \textsc{DynaEval} & \textsc{empathetic} & 0.17 &  0.10  &  -0.01 &  -0.02 \\
 \textsc{DEAM} & \textsc{DEAM} & \textsc{empathetic} & \bf 0.48 &  \bf 0.47  &  \bf 0.20 & \bf 0.20 \\
  \bottomrule 
\end{tabular}
\end{center}
\vspace{-0.5em}
\caption{Pairwise comparisons between DEAM with the proposed AMR-based manipulations and different baseline models using their original datasets, manipulations, and models (setup 2). All models have been trained on balanced sets of coherent/incoherent examples.}
\label{pair_model_comp}
\vspace{0.1em}
\end{table*}

\subsection{Manipulations Effect}

\begin{figure}[t]
\vspace{-1.5em}
\centering
\includegraphics[width=\linewidth]{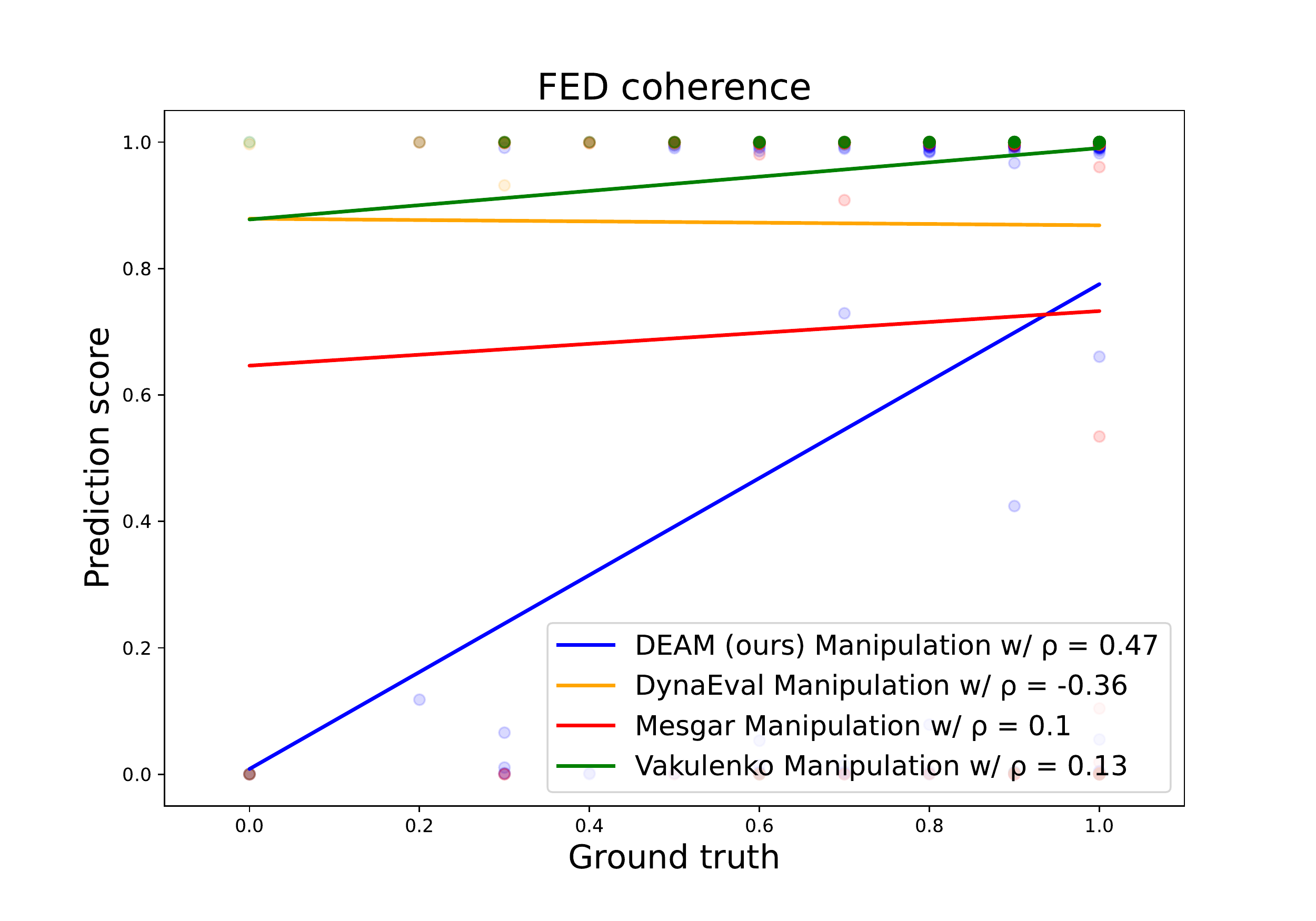}
\vspace{-1.5em}
\caption{Scatter plots and regression lines of different models predicted scores versus FED-coherence human evaluations. Overlapped points are represented darker.}
\label{regress_fed_coh}
\end{figure}

\begin{figure}[t]
\vspace{-1.5em}
\centering
\includegraphics[width=\linewidth]{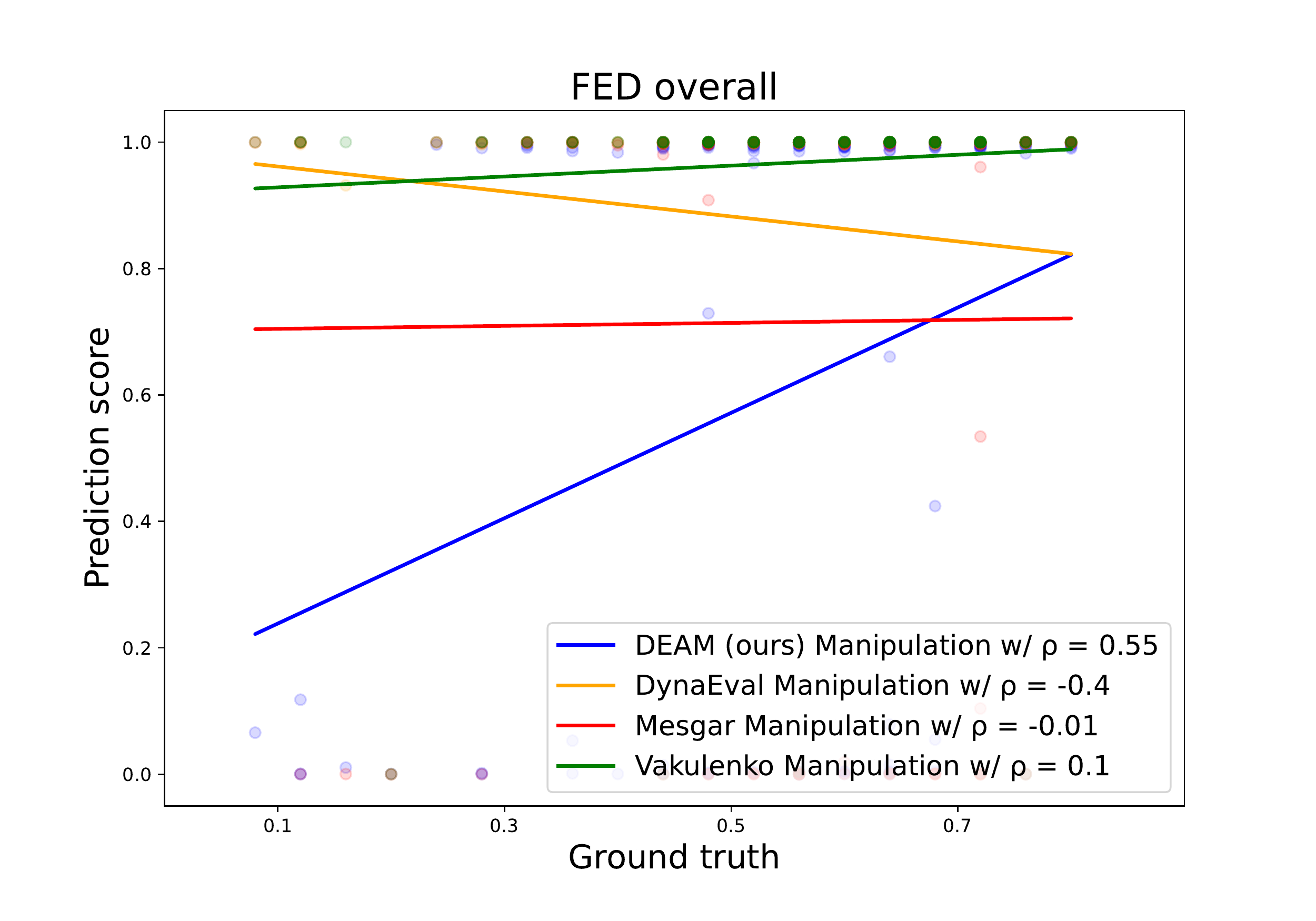}
\vspace{-2.5em}
\caption{Scatter plots and regression lines of different models predicted scores versus FED-overall human evaluations. Overlapped points are represented darker.}
\label{regress_fed_ovr}
\end{figure}

Table~\ref{mplts_comp_table} illustrates the results of experimental setup 3, where we fix both RoBERTa evaluator and \textsc{TopicalChat} and \textsc{PersonaChat} original datasets and apply different manipulation techniques to add negative samples. Even though the correlation for baseline manipulations increased drastically, which shows the effectiveness of strong pretrained language models in better encoding conversations information used for the evaluation task, DEAM's performance is still higher. This interprets the beneficial effect of AMR-based manipulations.  
The positive slops of the regression line in Figures ~\ref{regress_fed_coh} and ~\ref{regress_fed_ovr} between DEAM predicted coherence scores and human coherence and overall evaluations for FED dataset show the proposed manipulations superiority from a different angle. The distribution of baseline models predicted low scores for high-quality conversations and vice versa present their ineffectiveness in correctly distinguishing between low-quality and high-quality conversations. 

\subsection{Ablation Studies}

Next, we inspect the role of each of the four proposed manipulations in the metric's performance. We conduct an ablation study on \textsc{TopicalChat} and \textsc{PersonaChat} datasets to assess the effectiveness of each manipulation. For each specific manipulation, we remove it from the list of possible manipulations and try to randomly sample one up to three different manipulations to create negative samples. 

In Table~\ref{ablation}, we witness an overall drop by eliminating each of the manipulations that indicates the positive impact of all of the manipulations on generating higher quality negative samples that are closer to the samples generated by state-of-the-art models and consequently the evaluator's accuracy. Removing irrelevancy and decrease engagement manipulations have the most detrimental impact on the metric, which suggests that many state-of-the-art models struggle with such issues. By eliminating these manipulations the model does not have access to such negative examples during training, which significantly limits its ability to detect such incoherences during inference time. On the other hand, omitting coreference inconsistency from the manipulations has the lowest impact on DEAM, specifically for DSTC9 dataset which can be interpreted as the state-of-the-art models are safer regarding such issues. We also note that the performance difference between DSTC9 and FED could be due to the long conversations in DSTC9 that mostly include very limited coreferences. 

\begin{table}[t]
\vspace{-1em}
\begin{center}
\small
\begin{tabular}{lcccc}
\toprule
 \multirow{2}{*}{\textbf{Manipulation}} & \multicolumn{2}{c}{\bf FED} & \multicolumn{2}{c}{\bf DSTC9} \\ 
 & Coh & Overall & Coh & Overall \\
\midrule 
  \textsc{DEAM}
     & \bf 0.47 &  \bf 0.55  &  \bf 0.19 & \bf 0.20 \\
\midrule
  \textsc{-Contr}
   & 0.39 & 0.42 & 0.17 & 0.16\\ 
\midrule 
  \textsc{-CoRef\_Inconst.}
   & 0.41 & 0.46 & 0.19 & 0.20 \\ 
\midrule
  \textsc{-Irrel}  
    & 0.35  & 0.35  & 0.17 & 0.18 \\
\midrule
  \textsc{-Dec\_Eng}
     & 0.34 & 0.35  & 0.18 & 0.17 \\
  \bottomrule 
\end{tabular}
\end{center}
\caption{Correlations of DEAM with human judgments trained on different ablated manipulations.}
\label{ablation}
\vspace{-0.9em}
\end{table}

\begin{figure}[t]
\centering
\adjustbox{max width=\linewidth}{\includegraphics[width=0.72\linewidth]{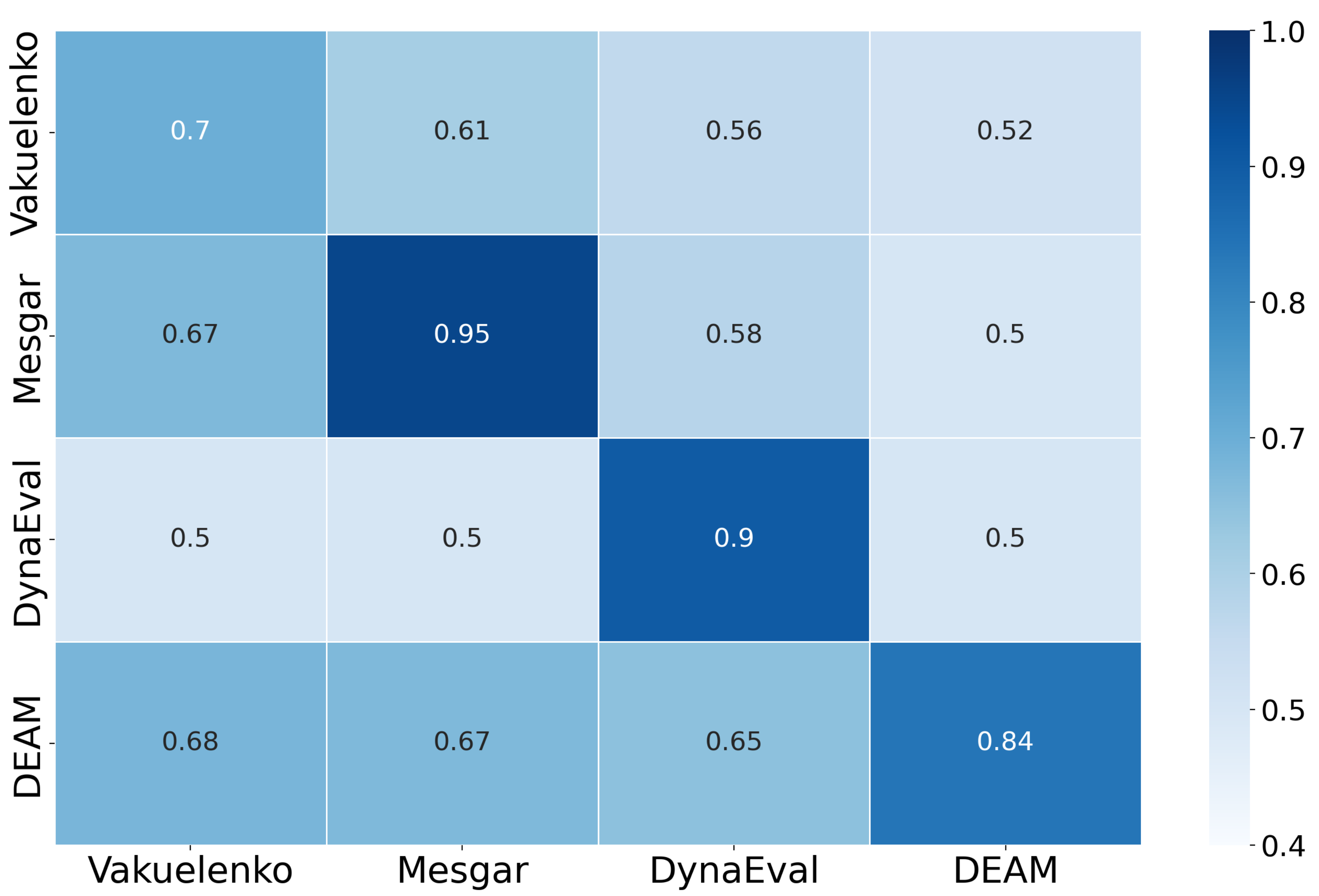}}
\caption{The accuracy of evaluation metrics to distinguish coherent/incoherent conversations in test data (y-axis) generated using baseline manipulations (x-axis).}
\label{cross_acc}
\vspace{-1.1em}
\end{figure}

\subsection{Qualitative Analysis}

We analyze the quality of DEAM versus baseline evaluators in terms of examining each model's performance to distinguish between positive and negative examples constructed leveraging various manipulations. Some examples are shown in Table \ref{mplts_examples} of Appendix. Figure~\ref{cross_acc} illustrates a heat map of the accuracy scores. X-axis and Y-axis show the manipulations used for creating training and testing datasets, respectively. As is expected the highest accuracies can be found on the diagonal where models have been trained and tested on datasets generated from pursuing the same manipulation techniques. The light-colored cells are mainly related to models trained on baseline manipulated data and tested on AMR-based perturbed data. This indicates that the baseline models trained on such types of text-level heuristic manipulations can not perform well and indeed have a random guess on more challenging incoherent examples that are generated by DEAM. While the higher accuracies of DEAM model on baseline test datasets show its capability to more effectively distinguish between positive and heuristically created their counterpart manipulations' negative conversations.

Our proposed manipulations in DEAM are directly influenced by the quality of Text-to-AMR and AMR-to-Text generation models. Even though finetuned T5~\mbox{~\cite{raffel2020t5}} models used here have been shown to be  effective~\mbox{~\cite{ribeiro2020amrtotext}}, there are still not perfect and suffer from some errors. We conducted a quick analysis of different deficiencies in conversations obtained by AMR-based back-translations. Most of these flaws are due to the fact that in Text-to-AMR generations some syntactic information such as verb tense, passive type of sentences, are removed from the text due to the semantic-based structure of the AMRs. Table{~\ref{analysis_back_translation}} in the Appendix shows such flaws. Ongoing work on improving AMR parsers and generators will lead to more robust AMR models, thus improving the quality of the proposed manipulations as well.

\begin{figure}[t]
\centering
\adjustbox{max width=\linewidth}{\includegraphics[width=0.75\linewidth]{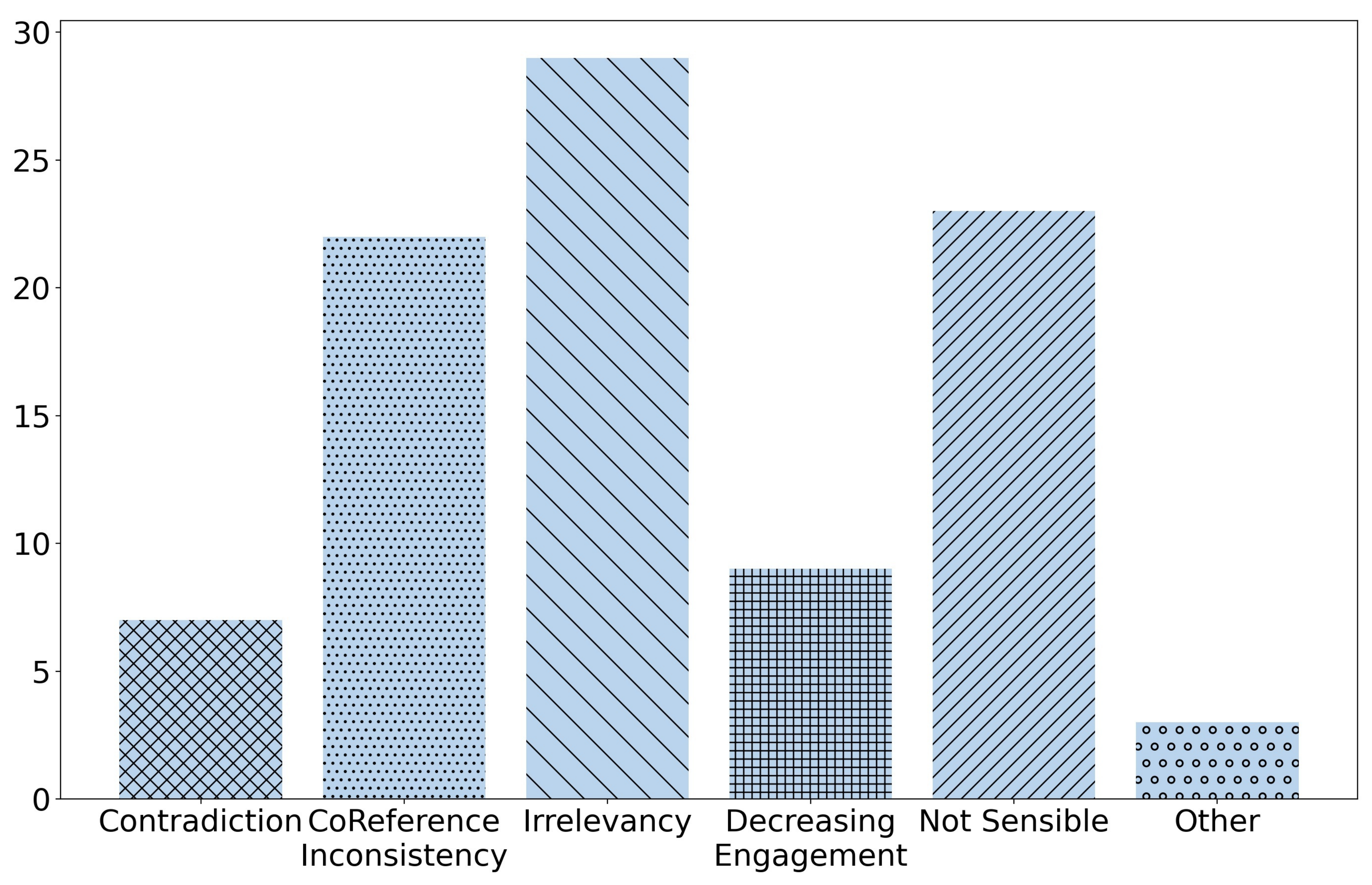}}
\caption{Statistics of different types of logical flaws observed in 50 randomly selected low-quality interactions between human and state-of-the-art dialogue systems (25 dialogues from FED and 25 from DSTC9)}
\label{error_dist}
\vspace{-1.2em}
\end{figure}

\subsection{Manipulation Coverage}
In the end, we conduct an analysis to explore the coverage rate of our proposed manipulations in the test datasets. To accomplish this, we analyze different commonly occurring logical flaws by advanced dialogue models via randomly selecting 25 low-quality interactions from FED~\cite{mehri2020fed} and 25 poor dialogues from DSTC9 datasets. The low-quality scores specified by human annotators indicate various types of flaws in the conversations. Our analysis suggests that we can classify those flaws into distinct categories as demonstrated in Figure{~\ref{error_dist}}. Most of the frequently happening flaws have been covered in our work except \textit{not\_sensibility} showing the sensibility of the generated responses. We leave \textit{not\_sensibility} evaluation for future works as replicating such issues besides the AMR-based manipulations mostly requires external knowledge bases which is not the focus of this work.

\section{Conclusion} \label{section:conclusion}

Reliable automatic trainable coherence evaluation metrics that can efficiently measure the dynamics of interactions between interlocutors are principally influenced by the quality of the training instances. We show that leveraging text-level manipulations can not adequately mirror the incoherence errors that current dialogue systems face. According to our study, DEAM can more effectively accomplish this task by relying on capabilities that AMR-based semantic perturbations and pretrained language models present. We leave the investigations regarding the effectiveness of AMRs for evaluating sensibility of the generated responses with taking into account knowledge bases for the future.

\section{Ethics} \label{ethics}

We acknowledge the importance of ACM Code of Ethics and totally agree with it. We ensure that our study is compatible with the provided code, specifically in the terms of providing non-offensive dataset construction.

In our proposed approach, we start from publically available human-human conversational datasets and attempt to apply our proposed manipulations at the AMR level. The main concern that arises here is the probability of generating offensive conversations from manipulated AMRs. The chance of such generations is faded with leveraging TopicalChat (Gopalakrishnan et al., 2019) and PersonaChat (Zhang et al., 2018) datasets that have been originally collected asking users to converse without profanity and inappropriate utterances. 
We should note that the possibility of perturbations that would have the possibility of generating objectionable outputs is not zero therefore we acknowledge there may be biases or abusive content via attacks that can be resolved by security trended studies which is out of this work’s scope.

\bibliography{anthology,custom}
\bibliographystyle{acl_natbib}
\appendix
\section{Appendix} \label{appendix}
In Table~\ref{mplts_examples}, we present examples of incoherent conversations generated by DEAM's AMR-based and other baselines' text-based manipulations each with a specific color. Due to the long length of the conversations, we include subsections from the conversations in that their utterances are separated by </UTT> separator. The colored lines and boldly written parts show types of manipulations and applied changes to the conversations using different approaches. It is obvious that the first three baseline manipulations result in very unnatural incoherent conversations while the last one which applies three out of four of our proposed semantic-based manipulations results in more challenging incoherent conversations that could be generated by state-of-the-art generative models. 

The original and manipulated AMR graphs of the conversation in Figure~\ref{manipul_plt_fig} is shown in Table~\ref{amr_exm}. We linearized the AMR graphs to be placed in the table.

Table~\ref{analysis_back_translation} demonstrates different syntactical issues that could be resulted from text-to-AMR and AMR-to-text models. The issues have been bolded in the table.

\begin{table*}[t]
  \centering
    \begin{tabularx}{\textwidth}{X}
    \toprule
      \fontsize{14}{8} \textbf{Incoherent Conversation}\\
    \midrule
    \bf\textcolor{blue}{Vakulenko et al., 2018 -- {Permute bolded section with another random dialogue}
    }\\
     ...</UTT>\textbf{No clue. Wonder if the Model T ever won an award? It is from 1908. It was one of the first cars that ws accessible to the masses.</UTT>What is your favorite car brand?</UTT>I like Cadillacs which were named after French explorer Antoine de la Mothe Cadillac who founded Detroit. It's about time for a drive now, goodbye!</UTT>1886 is thought of as the birth year of the modern car.</UTT>Yeah, I didn't know that. It was his Benz Patent-Motorwagen.</UTT>That is true, do you know when the first car was invented?}</UTT>Great question. I love that they experience eureka moments.</UTT>Yeah I never knew that,  thats pretty awesome.  Do you have a dog?</UTT>I do not have a dog. Do you?</UTT>I do,  shes 2,  a rescue.  I think my dog is the exception when they say dogs an elephants can understand pointing lol shes obivious.</UTT>Thats funny. Do you consider dogs to be man's best friend.</UTT>...\\
    \midrule
    \bf\textcolor{BurntOrange}{Mesgar et al., 2020 -- Shuffle all utterances} \\
     ...</UTT>Probably because its faster to get around on.  Oh and where do they keep their subway ticket?  Or do dogs ride free?</UTT>Yeah, I also heard he switched his limp on his leg the entire time filming and no one ever noticed!</UTT>I am confused how dogs in moscow use the subway.</UTT>Yes, I love dogs.</UTT>Hello,  do you like dogs?</UTT>Same here,  That is really impressive though,  but im not sure how they know which subway to take lol</UTT>I also do not know why they would need to use the subway.</UTT>Great question. I love that they experience eureka moments.</UTT>I do not have a dog. Do you?</UTT>My friends dogs know like 50 commands!  Even the command dance!  lol</UTT>That is amazing.</UTT>I did know that. We should teach them more than just simple comands like sit and paw.</UTT>Did you know dogs have 12 different blood types?</UTT>Thats funny. Do you consider dogs to be man's best friend.</UTT>Yeah thats weird lol it looks weird too.</UTT>Dogs drink with underside of their tongue!</UTT>...\\
    \midrule
    \bf\textcolor{ForestGreen}{Dynaeval-- Shuffle one speaker's utterances} \\
     ...</UTT>My friends dogs know like 50 commands!  Even the command dance!  lol</UTT>\textbf{Great question. I love that they experience eureka moments.}</UTT>Yeah thats weird lol it looks weird too.</UTT>\textbf{No, that is crazy! I wonder if they have their own version of O-.}</UTT>Same here,  That is really impressive though,  but im not sure how they know which subway to take lol</UTT>\textbf{Been great talking to you.}</UTT>Probably because its faster to get around on.  Oh and where do they keep their subway ticket?  Or do dogs ride free?</UTT>\textbf{Dogs drink with underside of their tongue!}</UTT>Yeah I never knew that,  thats pretty awesome.  Do you have a dog?</UTT>\textbf{I am confused how dogs in moscow use the subway.}</UTT>I do,  shes 2,  a rescue.  I think my dog is the exception when they say dogs an elephants can understand pointing lol shes obivious.</UTT>\textbf{Yes, I love dogs.}</UTT>... \\
    \midrule
    \bf \textcolor{plum}{DEAM -- \textsc{CoRef\_Inconst.}, \textsc{Contr.}, \textsc{Irrel.}} \\
    ...</UTT>My friend's dog knows like 50 commands. Even dancing to them. LOL!</UTT>The tongue dog is drinking from the underside!</UTT>LOL, that's weird, it looks weird too.</UTT>\textbf{I'm looking for ways you use the subway in Moscow.}</UTT>Same here. That's really impressive, but I'm not sure how they know which subway to take.</UTT>I also don't know why they need to use the subway.</UTT>probably because it gets around faster. And where do they keep their subway tickets? or free dog rides.</UTT>Great question. I love how they experience the eureka moment.</UTT>I never knew that. That was pretty awesome. Do you have a dog?</UTT>I don't have a dog, do you?</UTT>I did. She was rescued at age 2. I think my dog was the exception when they said, laughing out loud, "The dog and elephant can understand."</UTT>That's funny, \textbf{do you consider the dog on the subway my best friend? I don't have a dog, do you owe it?}</UTT>...\\
    \bottomrule
    \end{tabularx} 
  \caption{Examples of incoherent conversations resulted from different applied manipulation techniques}
  \label{mplts_examples}
\end{table*}

\begin{table*}[t]
  \centering
    \begin{tabularx}{\textwidth}{X}
    \toprule
      \fontsize{14}{8} \textbf{AMR graphs of a conversation}\\
    \midrule
    \textcolor{blue} {Have you watched Sesame Street?}
    \\
        (w / watch-01
              :ARG0 (y / you)
              :ARG1 (b / broadcast-program
                    :name (n / name
                          :op1 "Sesame"
                          :op2 "Street"))
              :polarity (a / amr-unknown))\\
        \textcolor{blue} {I used to when my kids were young. I liked Oscar the Grouch. He seemed realistic.}\\
        (m / multi-sentence
              :snt1 (u / use-02
                    :ARG0 (ii / i)
                    :time (y / young
                          :domain (p / person
                                :ARG0-of (h / have-rel-role-91
                                      :ARG1 ii
                                      :ARG2 (k / kid)))))
              :snt2 (l / like-01
                    :ARG0 (ii2 / i)
                    :ARG1 (p2 / person
                          :name (n / name
                                :op1 "Oscar"
                                :op2 "the"
                                :op3 "Ggrouch")))
              :snt3 (s / seem-01
                    :ARG1 (r / realistic-03
                          :ARG1 (h2 / he))))

        \textcolor{blue} {He was one of my favorite character as well, why is he green though? I've always wondered that.}\\
        (m / multi-sentence
              :snt1 (ii / include-91
                    :ARG1 (h / he)
                    :ARG2 (c / character
                          :ARG1-of (f / favor-01
                                :ARG0 (ii2 / i)))
                    :mod (a / as-well))
              :snt2 (h2 / have-concession-91
                    :ARG1 (g / green-02
                          :ARG1 (h3 / he)
                          :ARG1-of (c2 / cause-01
                                :ARG0 (a2 / amr-unknown))))
              :snt3 (w / wonder-01
                    :ARG0 (ii3 / i)
                    :ARG1 (t / that)
                    :time (a3 / always)))\\
        \textcolor{blue} {He was once orange though.}\\
        (h / have-concession-91
              :ARG1 (o / orange
                    :domain (h2 / he)
                    :time (o2 / once)))\\\hline
            \textcolor{blue} 
            {You listen to Sesame Street?}\\
        (w / listen-01
              :ARG0 (y / you)
              :ARG1 (b / broadcast-program
                    :name (n / name
                          :op1 "Sesame"
                          :op2 "Street"))
              :polarity (a / amr-unknown))\\
        \textcolor{blue} {I used to be when my kids were young. I like Oscar the Grouch. He seems realistic.}\\
        (m / multi-sentence
              :snt1 (u / use-02
                    :ARG0 (ii / i)
                    :time (y / young
                          :domain (p / person
                                :ARG0-of (h / have-rel-role-91
                                      :ARG1 ii
                                      :ARG2 (k / kid)))))
              :snt2 (l / like-01
                    :ARG0 (ii2 / i)
                    :ARG1 (p2 / person
                          :name (n / name
                                :op1 "Oscar"
                                :op2 "the"
                                :op3 "Grouch")))
              :snt3 (s / seem-01
                    :ARG1 (r / realistic-03
                          :ARG1 (h2 / he))))\\
        \textcolor{blue} {They are among my favorite characters as well. I always wonder that.}\\
        (m / multi-sentence
              :snt1 (ii / include-91
                    :ARG1 (h / they)
                    :ARG2 (c / character
                          :ARG1-of (f / favor-01
                                :ARG0 (ii2 / i)))
                    :mod (a / as-well))
              :snt2 (w / wonder-01
                    :ARG0 (ii3 / i)
                    :ARG1 (t / that)
                    :time (a3 / always)))\\
        \textcolor{blue} {He was orange once though, I used to be when my kids were young. I hate Oscar the Grouch, he doesn't seem realistic.}\\
        (m / multi-sentence
              :snt1 (h / have-concession-91
                    :ARG1 (o / orange
                          :domain (h2 / he)
                          :time (o2 / once))
              :snt2 (u / use-02
                    :ARG0 (ii / i)
                    :time (y / young
                          :domain (p / person
                                :ARG0-of (h / have-rel-role-91
                                      :ARG1 ii
                                      :ARG2 (k / kid)))))
              :snt3 (h / hate-01
                    :ARG0 (ii2 / i)
                    :ARG1 (p2 / person
                          :name (n / name
                                :op1 "Oscar"
                                :op2 "the"
                                :op3 "Grouch")))
              :snt4 (s / seem-01
                    :polarity -
                    :ARG1 (r / realistic-03
                          :ARG1 (h2 / he))))\\
    \bottomrule
    \end{tabularx} 
    \caption{Original (top) and manipulated (bottom) AMR graphs of the  conversation of Figure~\ref{manipul_plt_fig} }

  \label{amr_exm}
\end{table*}

\begin{table*}[t]
  \centering
    \begin{tabularx}{\textwidth}{X}
    \toprule
       \fontsize{14}{8} \textbf{ \textcolor{ForestGreen}{Text}--> AMR--> \textcolor{blue}{Text} Examples}\\
    \midrule
     \textcolor{ForestGreen}{Original Sentence: I do. Tim Duncan did not go to the NBA until he finished college.}\\
     (m / multi-sentence :snt1 (d / do-02 :ARG0 (ii / i)) :snt2 (g / go-02 :polarity :ARG0 (p / person :name (n / name :op1 "Tim" :op2 "Duncan")) :ARG4 (t / team :name (n2 / name :op1 "NBA")) :time (u / until :op1 (f / finish-01 :ARG0 p :ARG1 (c / college)))))\\
      \textcolor{blue}{Back-translated Sentence: I do. Tim Duncan \textbf{won't} go to the NBA until he \textbf{finishes} college.}\\\hline
     \textcolor{ForestGreen}{Original Sentence: Nice. He was really hated in 2012 when he decided to join MIami Heats}\\
     (m / multi-sentence :snt1 (n / nice-01) :snt2 (h / hate-01 :ARG1 (h2 / he)       :degree (r / really) :time (d / date-entity :year 2012 :time-of (d2 / decide-01   :ARG0 h2 :ARG1 (j / join-up-02  :ARG0 h2 :ARG1 (t / team :name (n2 / name :op1 "Miami" :op2 "Heats")))))))\\
      \textcolor{blue}{Back-translated Sentence: Nice. He \textbf{really hated} in 2012 when he decided to join the Miami Heats.}\\\hline
      \textcolor{ForestGreen}{Original Sentence: Yes the guy is set for life, trust me. Do you like LeBron?}\\
      (m / multi-sentence :snt1 (s / set-02 :ARG1 (g / guy) :ARG2 (l / life)       :ARG1-of (t / trust-01 :mode imperative :ARG0 (y / you) :ARG2 (ii / i))) :snt2 (l2 / like-01 :ARG0 y :ARG1 (p / person :name (n / name :op1 "LeBron"))\\
      \textcolor{blue}{Back-translated Sentence: Trust me, \textbf{guys are} set in life, \textbf{like LeBron?}}\\\hline
     \textcolor{ForestGreen}{Original Sentence: Wow is he not a huge part of the show though?}\\
      (h / have-concession-91 :ARG1 (h2 / have-part-91 :polarity - :ARG1 (s / show-04) :ARG2 (h3 / he) :mod (h4 / huge) :mod (w / wow :mode expressive)))\\
      \textcolor{blue}{Back-translated Sentence: Wow, \textbf{but he's not} a huge part of the show.}\\\hline
     \textcolor{ForestGreen}{Original Sentence: They use Blue Tube to share law enforcement videos.}\\
      ((u / use-01 :ARG0 (t / they):ARG1 (p / publication :name (n / name            :op1 "Blue" :op2 "Tube")):ARG2 (s / share-01 :ARG0 t :ARG1 (v / video :topic (e / enforce-01 :ARG1 (l / law)))))\\
      \textcolor{blue}{Back-translated Sentence: They \textbf{used} the Blue Tube to share \textbf{a video} about law enforcement.}\\\hline
     \textcolor{ForestGreen}{Original Sentence: you have to tame them they emulate the owner.}\\
      (o / obligate-01 :ARG1 (y / you) :ARG2 (t / tame-01 :ARG0 y :ARG1 (t2 / they :ARG0-of (e / emulate-01 :ARG1 (p / person :ARG0-of (o2 / own-01))))))\\
      \textcolor{blue}{Back-translated Sentence: You have to tame them \textbf{by emulating the owner}.}\\
    \bottomrule
    \end{tabularx} 
  \caption{Examples of original sentences, their linearized AMR graphs and back-translated sentences indicated with green, black and blue colors respectively. Bold parts of the sentences demonstrate the syntactical changes resulted from AMRs that usually do not cover such information.}
  
  \label{analysis_back_translation}
\end{table*}

\end{document}